\documentclass{ieeeaccess}
\usepackage{cite}
\usepackage{amsmath,amssymb,amsfonts}
\usepackage{algorithmic}
\usepackage{graphicx}
\usepackage{algorithm}
\usepackage{textcomp}

\usepackage{pgfplots}
\pgfplotsset{compat=1.18}
\usepackage{array}
\usepackage{multirow}
\usepackage{booktabs}
\usepackage{balance}
\usepackage{stfloats}
\usepackage{dsfont}

\usepackage{bm}
\usepackage{siunitx}

\makeatletter
\AtBeginDocument{\DeclareMathVersion{bold}
\SetSymbolFont{operators}{bold}{T1}{times}{b}{n}
\SetSymbolFont{NewLetters}{bold}{T1}{times}{b}{it}
\SetMathAlphabet{\mathrm}{bold}{T1}{times}{b}{n}
\SetMathAlphabet{\mathit}{bold}{T1}{times}{b}{it}
\SetMathAlphabet{\mathbf}{bold}{T1}{times}{b}{n}
\SetMathAlphabet{\mathtt}{bold}{OT1}{pcr}{b}{n}
\SetSymbolFont{symbols}{bold}{OMS}{cmsy}{b}{n}
\renewcommand\boldmath{\@nomath\boldmath\mathversion{bold}}}
\makeatother

\def\BibTeX{{\rm B\kern-.05em{\sc i\kern-.025em b}\kern-.08em
    T\kern-.1667em\lower.7ex\hbox{E}\kern-.125emX}}

\NewSpotColorSpace{PANTONE}
\AddSpotColor{PANTONE} {PANTONE3015C} {PANTONE\SpotSpace 3015\SpotSpace C} {1 0.3 0 0.2}
\SetPageColorSpace{PANTONE}%
  
\definecolor{tab:gray}{rgb}{0.5,0.5,0.5}
\definecolor{tab:blue}{rgb}{0.1216,0.4667,0.7059}
\definecolor{tab:red}{rgb}{0.8392,0.1529,0.1569}

\begin{document}
\history{Preprint. Accepted for publication in IEEE Access, July 2026. The final published article will be available in IEEEXplore.}
\doi{}

\title{Learning to Walk With Less: A Dyna-Style Approach to Quadrupedal Locomotion}
\author{
    \uppercase{Francisco Affonso}\authorrefmark{1, 2},
    \uppercase{Felipe Tommaselli}\authorrefmark{1},
    \uppercase{João H. Aléssio}\authorrefmark{1}, 
    \uppercase{Vivian S. Medeiros}\authorrefmark{1}, \\ 
    \uppercase{Mateus V. Gasparino}\authorrefmark{2}, 
    \uppercase{Girish Chowdhary}\authorrefmark{2}, 
    \uppercase{Marcelo Becker}\authorrefmark{1}}

\address[1]{Mobile Robotics Group, São Carlos School of Engineering, University of São Paulo (USP), BR.}
\address[2]{Field Robotics Engineering and Science Hub (FRESH), University of Illinois at Urbana-Champaign (UIUC), USA}

\tfootnote{The publication was written prior to Mateus V. Gasparino joining Amazon.}

\markboth
{F. Affonso \headeretal: Learning to Walk With Less}
{F. Affonso \headeretal: Learning to Walk With Less}

\corresp{Corresponding author: Francisco Affonso (faffonso@illinois.edu).}

\begin{abstract}
Traditional on-policy reinforcement learning (RL) controllers for quadrupedal locomotion often suffer from low data efficiency, requiring millions of interactions with simulated environments to achieve stable control. We integrate model-based techniques that improve sample efficiency by augmenting PPO rollouts with synthetic data in a Dyna-style framework. Our method employs a learned transition model to generate short-horizon synthetic tails for each trajectory, anchored by physics-based simulation to preserve stability. A predefined scheduling strategy gradually integrates synthetic transitions, preventing model usage during early training stages when prediction accuracy is low. Through extensive ablation studies, we analyze how varying data parameters influence PPO’s learning behavior. Finally, we validate our method in simulation on a Unitree Go1 robot, reaching convergence with substantially fewer simulation steps (19.64M vs. 27.53M) and a 12.24\% reduction in wall-clock training time, without compromising policy performance or convergence. Cross-platform experiments on ANYmal and Unitree Go2 further confirm the framework's ability to learn high-dimensional locomotion control with substantially reduced simulation experience, despite reward trade-offs on complex morphologies.
\end{abstract}

\begin{keywords}
Legged robots, reinforcement learning, data efficiency.
\end{keywords}

\titlepgskip=-21pt

\maketitle

\section{Introduction}

\IEEEPARstart{L}{ocomotion} control of quadrupedal robots is the coordination of movements and interactions with the environment to achieve desired motion and maintain stability.  Beyond the need to move the body in the desired direction according to a given command, it is standard practice to provide a gait pattern due to the complexity of online handling foot-ground interactions. To address this challenge, traditional methods have relied heavily on trajectory optimization (TO) and model predictive control (MPC) techniques~\cite{grandia2023perceptive, cebe2021online}, which optimize the robot's motion by leveraging constraints and kinodynamic models. Additionally, gait generation methods provide essential information to the controller, specifying which gait pattern (e.g., trot or walk) should be followed. 

Although recent works in TO and MPC have extended these methods to handle whole-body dynamics and even contact-implicit formulations without pre-defined gait sequences~\cite{le2024fast, amatucci2024accelerating, mastalli2023inverse}, they still face limitations. In general, these approaches struggle with discontinuous rewards, which necessitate heavily pre-processed terrain information, and gait optimization can often converge to local minima. 

\newpage

Given these challenges, deep reinforcement learning (RL) emerges as a promising alternative. By leveraging a trained policy, RL treats locomotion as a continuous control problem, mapping current states to actions without the need for computationally expensive real-time planning. Furthermore, RL policies can account for long-term objectives and accommodate environmental uncertainties, such as friction variations or terrain non-linearities, that frequently degrade the performance of traditional model-based frameworks.

These policies learn suitable actions through a trial-and-error approach during the agent's interaction with real or simulated environments, where rewards quantify how accurately the motion follows the desired command and style. However, RL requires extensive data to explore the state space and identify the best policy. This is particularly evident in on-policy settings, where the same policy must be both evaluated and updated. This dependency requires new trajectory rollouts for every iteration, significantly hindering sample efficiency. This challenge is evident in recent RL-based locomotion controllers, which necessitate thousands of hours of simulated experience across parallelized environments to achieve convergence and stabilize control policies~\cite{haarnoja2019learning, hwangbo2019learning}.

On-policy methods are widely preferred because they ensure gradient updates strictly reflect the agent’s current behavior, leading to stable and continuous policy improvements~\cite{margolis2023walk, levy2024learning}. However, this stability comes at the cost of high environmental interaction. Given these trade-offs, it is critical to investigate methods that enhance sample efficiency without compromising its robust convergence properties.

In this context, most of the on-policy frameworks rely on model-free reinforcement learning (MFRL), i.e., where the agent optimizes its behavior without an internal representation of the environment's transition dynamics. To mitigate the resulting data bottleneck, model-based reinforcement learning (MBRL) methods incorporate known or learned models to enhance policy training. In this context, neural networks are commonly employed as function approximators to model the complexity and uncertainty of robotic motion~\cite{chua2018deep, Hansen2022tdmpc, krinner2025accelerating}.

A key method for incorporating predictive models into an MBRL framework is through data augmentation, as outlined by the Dyna-style paradigm~\cite{sutton1991dyna}. Instead of integrating the model directly into the policy’s decision-making process, Dyna-style frameworks generate synthetic training data to supplement the conventional MFRL training process~\cite{janner2019trust, dong2024dyna}. This strategy not only leverages the robustness of MFRL-based policies but also enhances sample efficiency through simulated experience derived from the learned transition model. However, the central challenge in integrating this paradigm into on-policy methods lies in effectively combining synthetic and simulated data while mitigating the inevitable distribution shifts that arise between the predictive model and the actual environment.

In this paper, we introduce a Dyna-style MBRL framework that improves sample efficiency for PPO-based quadrupedal locomotion controllers by progressively substituting trajectory tails with short-horizon synthetic data from a learned transition model. By confining synthetic transitions to trajectory tails, each new rollout re-initializes from the final ground-truth simulation state, preventing compounding transition errors across episodes. Through systematic ablation studies, we identify optimal data augmentation parameters and demonstrate that our scheduled approach reduces wall-clock training time by 12.24\% on Unitree Go1 while achieving superior or comparable maximum rewards and policy stability compared to vanilla PPO baselines.

The key contributions of this work are:
\begin{itemize} 
    \item Integrating a Dyna-style MBRL framework that enhances sample efficiency in on-policy settings by substituting short-horizon synthetic trajectory tails into PPO rollouts, utilizing a ground-truth anchoring mechanism to prevent compounding errors;
    \item An ablation study under vanilla conditions evaluating how the number of parallel environments and rollout horizons affect the training of locomotion controllers;
    \item Cross-platform validation demonstrating the proposed method's effectiveness across Unitree Go1/Go2 and ANYmal-C/D platforms.
\end{itemize}

\newpage
\section{Related Work}

\emph{Reinforcement Learning} (RL) has emerged as a compelling alternative to model-based optimal control approaches. Unlike these model-based controllers, model-free RL methods do not require an explicit system model; instead, they learn policies directly through trial-and-error interactions with real or simulated environments, mapping appropriate actions to each state. Miki et al.~\cite{miki2022learning} and Kumar et al.~\cite{kumar2021rma} demonstrate that policies can learn to locomote in diverse real-world scenarios and adapt to various unstructured environments, provided that the training process is robust enough to expose the agent to a wide range of conditions (e.g., locomotion commands, scenarios, and initial positions) and employs reward functions aligned with the desired locomotion behavior.

Due to the need for extensive exploration across a wide state space and exposure to diverse environments and scenarios, several techniques have been applied to enhance the RL training process for locomotion controllers. Margolis et al.~\cite{margolis2024rapid} applied adaptive curriculum learning, a method that allows the agent to gradually improve by initially training on simple locomotion tasks before progressively increasing task complexity as learning advances. This structured approach prevents the agent from being overwhelmed by highly challenging tasks at the beginning of training and enables a more stable and efficient learning process. Additionally, Rudin et al.~\cite{rudin2022learning} introduce the use of massively parallel platforms to accelerate data collection, significantly increasing the amount of experience available for training policies. However, while these techniques improve policy learning, they do not directly address the challenge of more efficiently utilizing real-world or simulated samples during training.

\emph{Model-Based Reinforcement Learning} (MBRL) enhances standard RL by incorporating a transition model to guide the learning process. This model enables the agent to perform online planning through trajectory rollouts~\cite{zhao2025learning} or to improve sample efficiency via synthetic data generation~\cite{hoffman2025learning}. Yang et al.~\cite{yang2020data} introduced initial designs that integrate a learned model to generate sampled trajectories, following the principles established in Chua et al.~\cite{chua2018deep}. While this approach reduces the training iterations relative to model-free baselines, its performance as an online sampling-based planning method is strictly contingent upon the fidelity of the learned model; consequently, inaccurate predictions directly propagate to suboptimal control behaviors.

Song et al.~\cite{song2024learning} propose an MBRL method that retains the benefits of MFRL while addressing sample efficiency without incorporating the model into the online decision-making process. Their approach uses a dynamic representation of legged robots that enables differentiable simulation during training. This method facilitates faster convergence and more stable training by computing low-variance first-order gradients based on robot dynamics, thereby improving overall learning efficiency. However, it does not investigate the incorporation of transition models for data augmentation in on-policy settings, following the Dyna-style paradigm~\cite{sutton1991dyna}.

\newpage

In parallel, there exist data augmentation methods that leverage the inherent symmetry of quadrupedal robots by flipping data across a symmetric plane or axis~\cite{su2024leveraging, hoeller2024anymal}. Although these techniques are mainly applied to complex tasks such as loco-manipulation or locomotion through challenging terrains, they do not interfere with Dyna-style frameworks, allowing both approaches to be used together.

Despite progress in Dyna-style MBRL~\cite{dong2024dyna}, its integration with on-policy frameworks remains under-explored, particularly for high-dimensional locomotion tasks. The primary challenge lies in the management of mixed-data buffers, where transitions from physics-based simulations are augmented by synthetic model rollouts. Unlike off-policy methods that can tolerate high-variance data from diverse sources, on-policy training requires rigorous alignment between the model's transition distribution and the agent's current state-action manifold. Without this synchronization, synthetic data can introduce modeling biases that lead to distribution shifts, ultimately destabilizing the training process and causing policy divergence.
\section{Method}

\begin{figure*}[t]
    \centering
    \includegraphics[width=0.95\linewidth]{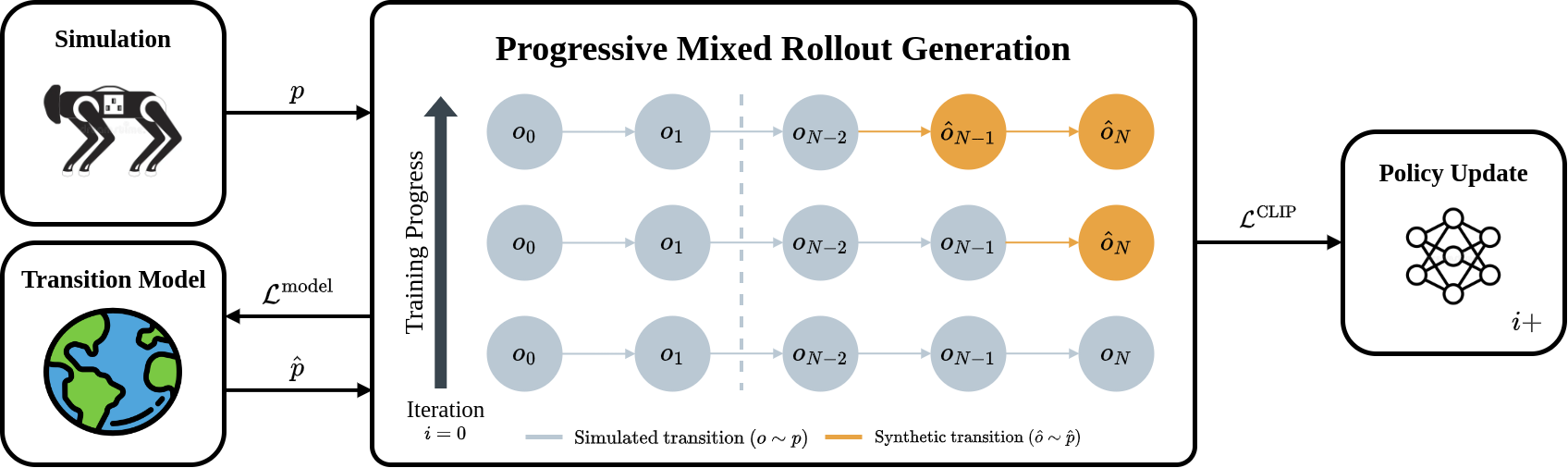}
    \caption{Overview of the proposed MBRL-based locomotion controller training framework. The pipeline integrates physics-based simulation with a learned transition model to generate progressive mixed rollouts to improve sample efficiency. Initially, the agent collects entirely ground-truth transitions (gray) from the simulator $p$. As training progresses, the transition model $f_\theta$ is optimized using the rollout buffer to minimize $\mathcal{L}_{\text{model}}$. This model is then used to substitute a growing portion of the trajectory tail with synthetic transitions (orange), $\hat{o} \sim \hat{p}$, allowing the agent to learn faster by minimizing the need for expensive simulation data. The resulting mixed trajectories are utilized to compute the policy loss ($\mathcal{L}_{\text{CLIP}}$) for the on-policy update. By anchoring the start of each new iteration in the last verified ground-truth state, the framework maintains the temporal consistency required for stable locomotion while significantly reducing the total number of environment interactions required for convergence.}
    \label{fig:pipeline_overview}
\end{figure*}

In this section, we present our method to improve data efficiency in training PPO-based reinforcement learning locomotion controllers. As illustrated in Fig.~\ref{fig:pipeline_overview}, rather than extending trajectories, we employ a substitution strategy where the total number of steps per rollout remains invariant. As training progresses, the later portions of each simulated trajectory are increasingly replaced by synthetic transitions generated by the learned model, governed by a predefined scheduler. This approach reduces the reliance on physics simulations while maintaining the consistent-length sequences required for stable PPO updates. Furthermore, by confining synthetic data to the trajectory tail, we ensure that model inaccuracies do not propagate across rollout boundaries, keeping the initial conditions of each iteration grounded in the simulation data.

\subsection{Problem Formulation}

We formulate the continuous control problem in discrete time at each timestep $k$. Given the absence of exteroceptive sensors, the system is treated as a Partially Observable Markov Decision Process (POMDP). However, since training is restricted to flat terrains, we assume that the observation $o_k$ provides a sufficiently informative representation to be treated as a Markovian state $s_k = o_k$.

With that, we define the tuple  $(S, A, p, r)$ that specifies our problem, where $S$ is the set of possible states, $A$ is the set of possible actions, $p(o_{k+1} | o_k, a_k)$ represents the transition function, and $r(o_k, a_k)$ defines the reward function. The goal of RL is to find the best policy $\pi^*$ that maximizes the expected discounted sum of rewards, using a discount factor $\gamma \in [0, 1)$, over the configuration defined by the tuple:

\begin{equation}
\pi^* = \arg\max_{\pi} \mathbb{E}_{\pi} \left[ \sum_{k=0}^{\infty} \gamma^k \cdot r(o_k, a_k) \right].
\end{equation}

The optimization of $\pi$ is performed using PPO, an on-policy algorithm that employs a clipped surrogate objective to ensure that the policy update remains within a trust region~\cite{schulman2015trust}, preventing the updated policy from deviating excessively from the behavior policy used for data collection.

The objective function is defined as:
\begin{equation}
        \mathcal{L}^{\text{CLIP}}(\phi) = \mathbb{E}\left[ \min ( \rho(\phi) \hat{A},    \text{clip}(\rho(\phi), 1-\epsilon, 1+\epsilon) \hat{A} )\right],
\label{eq:clip}
\end{equation}
where $\hat{A}$ is the estimated advantage, and $\rho(\phi)$ measures the deviation between the current policy $\pi_\phi$ and the old policy $\pi_{\text{old}}$ used during data collection. The hyperparameter $\epsilon$ bounds policy updates to ensure training stability.

In standard PPO, the transition distribution $p(o_{k+1}|o_k, a_k)$ is provided exclusively by the environment. By introducing a learned transition model $\hat{p}(\hat{o}_{k+1}|o_k, a_k)$, we introduce a distribution shift in the trajectory rollouts where synthetic predictions are utilized.

While this shift can facilitate broader exploration of the state space and reduce the computational cost of simulated steps, it poses a risk to the trust-region assumption. Specifically, if the model distribution $\hat{p}$ diverges significantly from the ground-truth environment distribution $p$, the estimated advantages may become biased. This mismatch can lead to inaccurate gradient directions and unstable policy updates, potentially violating the stability guarantees that make PPO effective for locomotion. The following sections describe how we address this challenge while preserving the sample efficiency benefits of model-based augmentation.

\subsection{State and Action Space}

To effectively implement the data augmentation process described in Section~\ref{subsec:data-augmentation}, it is essential to design the state and action spaces to facilitate accurate predictions using an approximate transition model. This entails ensuring that the current state encapsulates information that can be approximated from the preceding state and action. We incorporate both direct and indirect proprioceptive data into our state representation, an approach that ensures consistent modeling across diverse legged robots. 

As the policy must track locomotion commands while maintaining temporal context, the input state is augmented with velocity commands ${c}_k = [v_x^{\text{cmd}}, v_y^{\text{cmd}}, \omega_z^{\text{cmd}}]$ and the previous action ${a}_{k-1}$. It is important to note that these components are not transitions of the environment, but rather exogenous or policy-dependent variables. Consequently, we partition the observation $o_k$ into a predictable proprioceptive vector $o_k^{prop}$ and a non-predictable context vector $o_k^{ctx}$:
\begin{align}
    &s_k =  o_k = [o_k^{\text{prop}}, o_k^{\text{ctx}}], \nonumber \\
    o_k^{prop} = [{q_k}, &\dot{{q_k}}, {g_k}, {v_k}, {\omega_k}], \quad o_k^{ctx} = [{c}_k, {a}_{k-1}],
\end{align}
where $q, \dot{q} \in \mathbb{R}^{12}$ represent the joint positions and velocities, $g \in \mathbb{R}^{3}$ is the projected gravity, and $v, \omega \in \mathbb{R}^{3}$  correspond to the linear and angular velocities of the base, respectively.

Regarding the actions, they are defined as the target positions for each joint (${A}\in \mathbb{R}^{12}$), which are tracked by a PD controller to achieve the desired positions.

\subsection{Rewards}
\label{sec:rewards}

To reward the robot's movement, two groups of rewards are used: $r^{\text{task}}$ to reward the main task of the policy, i.e., tracking the locomotion commands, and $r^{\text{style}}$ to penalize undesirable behaviors (e.g., excessive vertical velocity or unsmooth joint actions) and regularize the action space to ensure hardware-feasible movements. The final reward is computed as the weighted sum of these components, with specific reward formulations and coefficients detailed in Appendix~\ref{app:additional_params} for each experimental platform.

Notably, our formulation does not include components to enforce specific gait patterns or contact sequences. Consequently, the learning process is unconstrained, allowing gait patterns to emerge organically as the optimal solution for the specified velocity tracking and regularization objectives.

\subsection{Data Augmentation}
\label{subsec:data-augmentation}

Recognizing the massive data requirements of on-policy RL and the computational overhead of high-fidelity physics simulations for computing transitions $p(o_{k+1} | o_k, a_k)$,  this work introduces a learned dynamics model to optimize data collection. Our approach progressively replaces the later segments of simulated trajectories with synthetic rollouts, maintaining a constant trajectory length $N$ while reducing the simulation steps required per policy update. The execution flow of this hybrid data generation process is detailed in Algorithm~\ref{alg:rl_algorithm}.

\begin{algorithm}[t]
\caption{Proposed MBRL Training Process}
\label{alg:rl_algorithm}
\begin{algorithmic}[1]
\STATE \textbf{Initialize:} policy $\pi_{\phi}$, predictive model $f_\theta$, empty replay buffer $\mathcal{B}$, and total rollout length $N$.
\FOR{$i = 1$ to $I$}
    \STATE \textbf{Determine} the synthetic rollout length using the scheduler and reset buffer:
    \STATE \hspace{0.5cm} $N_s \gets g(i)$ \hfill {$\triangleright$ See equation~(\ref{eq:schedular})}
    \STATE \hspace{0.5cm} $N_r \gets N - N_s$
    \STATE \hspace{0.5cm} $\mathcal{B} \gets \emptyset$ \hfill $\triangleright$ Flush buffer
    \STATE \textbf{Collect} $N_r$ steps using $\pi_\phi$ in the simulated environment:
        \STATE \hspace{0.5cm} $a_k \gets \pi_\phi(o_k)$
        \STATE \hspace{0.5cm} $(o_{k+1}, r_k) \gets p(o_k, a_k)$
        \STATE \hspace{0.5cm} $\mathcal{B} \gets \mathcal{B} \cup (o_k, a_k, o_{k+1}, r_k)$
    \STATE \textbf{Train} the predictive model on $\mathcal{B}$:
        \STATE \hspace{0.5cm} $\mathcal{L}^{\text{model}}(\theta) =  \text{Huber}(\mathcal{B}; f_\theta)$ \hfill {$\triangleright$ See equation~(\ref{eq:model_loss})}
        \STATE \hspace{0.5cm} $\theta \gets \text{update}(\mathcal{L}^{\text{model}}(\theta))$
    \STATE \textbf{Generate} $N_s$ synthetic steps for data augmentation:
        \STATE \hspace{0.5cm} $a_k \gets \pi_\phi(o_k)$
        \STATE \hspace{0.5cm} $( \hat{o}_{k+1}, \hat{r}_k) \gets  \hat{p}(o_k, a_k; f_\theta)$
        \STATE \hspace{0.5cm} $\mathcal{B} \gets \mathcal{B} \cup  (o_k, a_k, \hat{o}_{k+1}, \hat{r}_k)$
    \STATE \textbf{Update} the policy:
    \STATE \hspace{0.5cm} $\mathcal{L}^\text{CLIP} (\phi) \gets \text{PPO}(\mathcal{B})$  \hfill {$\triangleright$ See~\cite{schulman2017proximal}}
    \STATE \hspace{0.5cm} $\phi \gets \text{update}(\mathcal{L}^\text{CLIP} (\phi) )$ 
\ENDFOR
\end{algorithmic}
\end{algorithm}

In this work, the learned transition model is implemented as a multilayer perceptron (MLP) that leverages the universal function approximation capabilities of neural networks as a deterministic predictor~\cite{hornik1989multilayer}. Empirically, this is sufficient in our setting, where flat-terrain locomotion exhibits short horizons and low-variance dynamics.

Moreover, our model predicts only the proprioceptive state sequence $o^\text{prop}$. The non-predictable components are computed from the current state and prior action, then concatenated with the model's proprioceptive predictions to form the complete transition. Furthermore, because our reward depends on quantities beyond the observation (e.g., torque, joint acceleration), the model must also predict the scalar reward $\hat{r}_k$. The transition mapping is defined as:
\begin{equation}
    \hat{p}(\hat{o}_{k+1} \mid o_k, a_k; {f_\theta}), \quad \hat{o}_{k+1}^{\text{prop}}, \hat{r}_k = f_\theta(o_k, a_k),
\end{equation}
where $\theta$ denotes the network's weight parameters and the superscript $\hat{}$ an estimated variable.

By predicting the reward alongside the next state, the model can fully replace the simulation step. Moreover, since the policy outputs the next action for each state ($a_k = \pi_\phi(o_k)$), it is possible to recursively predict an entire rollout using the predictive model, starting from an initial state.

We consider a rollout as a sequence of $N$ state-action pairs, each accompanied by its corresponding reward and next state, stored as $( o_k, a_k, o_{k+1}, r_k )$ in a replay buffer. These rollouts can be generated either through the interaction of the policy with the simulation, which produces simulated data ($N_r$), or through the interaction with the predictive model, which produces synthetic data ($N_s$). The final rollout maintains a constant total length, $N = N_r + N_s$, with only the proportion of simulated to synthetic data varying. This combined dataset is then leveraged in the reinforcement learning update algorithm.

Building on the explanation of the predictive model, we now present how traditional RL-based locomotion controllers can integrate the concept of data augmentation.

The process begins by initializing the rollout buffer $\mathcal{B}$ with $N_r$ transitions collected directly from the physics simulation $p(\cdot)$. These ground-truth samples are used to optimize the transition model $f_\theta$. The model is trained by minimizing the Huber loss between the predicted proprioceptive state and reward and their corresponding ground-truth values across $R$ parallel robots:

\begin{equation}
    \mathcal{L}^{\text{model}}(\theta) = \sum_{r=1}^{R} \sum_{k=1}^{N_r} L_{\delta} \left( (o^{\text{prop}}_{k+1},\, r_k ) - f_\theta({o}_k, a_k) \right),
    \label{eq:model_loss}
\end{equation}
where the Huber operator $L_{\delta}$ acts as a squared norm error when below $\delta$ and as a linear error otherwise.

\begin{figure}[t]
    \centering
    \includegraphics[width=1.0\linewidth]{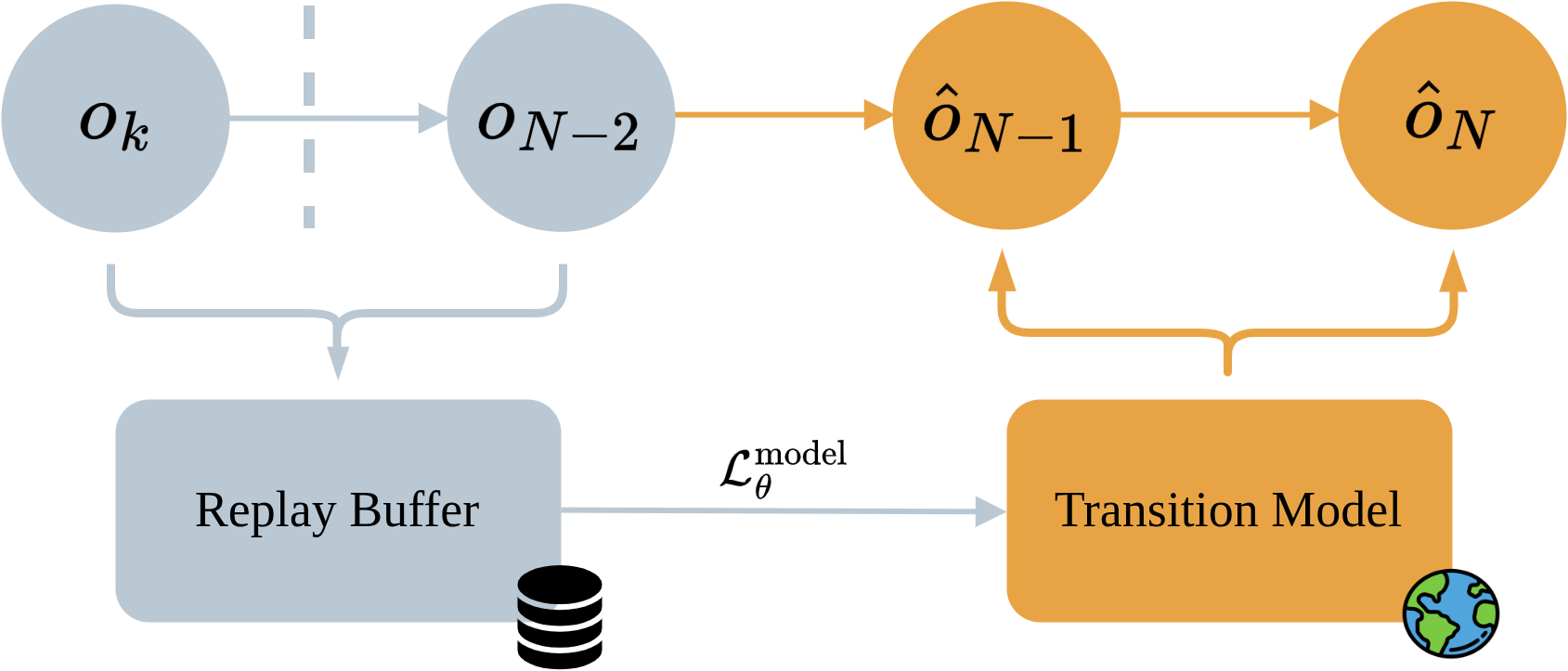}
    \caption{Visualization of the hybrid rollout generation and transition model training. Simulated observations ($o_k \dots o_{N-2}$) are stored in a rollout buffer to minimize the model loss $\mathcal{L}^\text{model}(\theta)$. By predicting both the next state and the corresponding reward, the model replaces simulation steps. The policy recursively interacts with this model to generate synthetic rollouts.}
    \label{fig:model_training}
\end{figure}

With the model trained using data generated from the interaction between the policy and the environment, it abstracts essential relationships at the current stage of the robot's locomotion. To maintain synchronization with the evolving policy, the model is re-optimized immediately following each simulation phase. This ensures synthetic transitions remain consistent with the ground-truth dynamics observed in that iteration, as illustrated in Fig.~\ref{fig:model_training}.

However, since the model must achieve high accuracy to generate useful data for the learning policy algorithm, we gradually integrate synthetic rollouts by varying the amount of synthetic data used during training:

\begin{equation}
    N_s = g(i) = \text{min} \left( \text{max} \left( x + \frac{i - a }{b - a} \cdot (y - x), x\right), y \right)
    \label{eq:schedular},
\end{equation}
where $i$ represents the current iteration within the learning process and $a$ and $b$ are the start and end iterations over which the synthetic rollout length $N_s$ ramps linearly from $x$ to $y$, with $y$ representing the maximum rollout length extension.

The scheduler, formulated in equation~(\ref{eq:schedular}), progressively increases the number of synthetic steps as training iterations progress while decreasing the number of simulated steps to maintain a consistent rollout length throughout the training process. This progression, illustrated in Fig.~\ref{fig:pipeline_overview}, allows the agent to rely more heavily on the transition model as training maturity increases. Notably, we employ a predefined curriculum rather than an adaptive metric (e.g., prediction error) to govern synthetic rollout length. This approach eliminates the computational instability and overhead of online accuracy monitoring, ensuring a strictly bounded increase in synthetic transitions as training progresses.

\begin{figure}[b]
    \centering
    \includegraphics[width=1.0\linewidth]{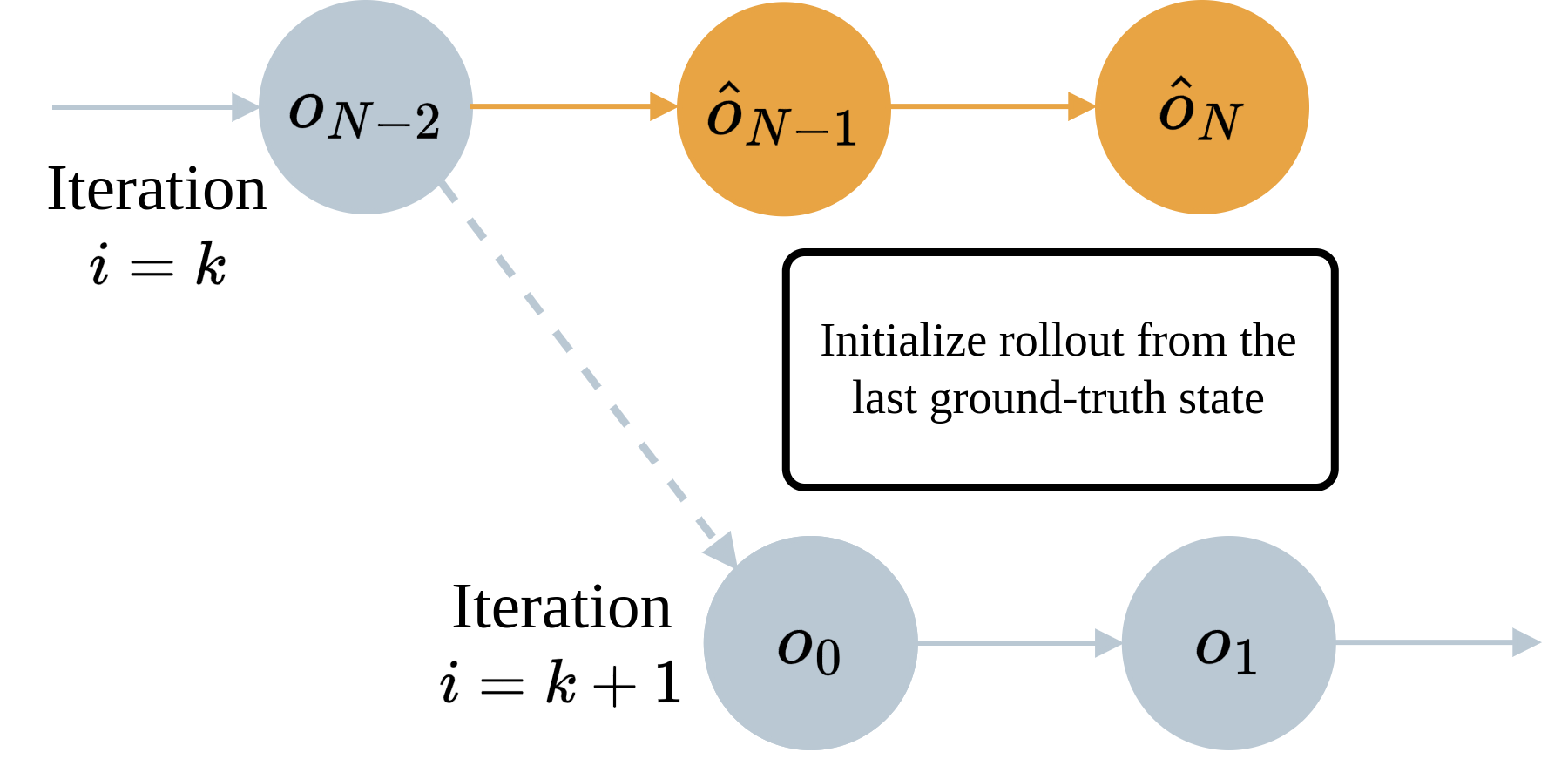}
    \caption{Ground-truth anchoring and reset mechanism. To ensure temporal stability, synthetic transitions (orange) are confined to the trajectory tail. For the subsequent iteration, the rollout re-initializes from the terminal ground-truth state instead of the predicted state. This reset mechanism grounds initial conditions in the true dynamics $p(\cdot)$, preventing compounding model errors across training episodes.}
    \label{fig:tail}
\end{figure}

In addition, synthetic transitions are exclusively appended to the trajectory tail to prevent model inaccuracies from affecting the integrity of subsequent rollouts. By terminating the rollout with synthetic steps and resetting the buffer to the last ground-truth state, the next iteration always begins from a state grounded in the true dynamics $p(\cdot)$ as illustrated in Fig.~\ref{fig:tail}. This prevents the compounding error of the transition model from causing the agent's initial conditions to diverge in subsequent data collection cycles.

Furthermore, due to the predictive model's limited accuracy over long horizons, short-horizon synthetic sampling is employed. Moreover, although branching augmentation is common in off-policy RL to increase sample diversity, it is ill-suited to on-policy frameworks like PPO, which require continuous temporal sequences for reliable advantage and value estimates. 

Rather than fragmenting rollouts into disjointed segments, our approach uses a single linear extension that preserves the chronological flow, letting the agent capture long-term returns and maintain a stable value function. However, it is important to be aware that, because termination events such as collisions are not predicted, tasks with frequent early termination may produce substituted tails that contain physically invalid transitions; such cases are outside the scope of the flat-terrain locomotion setting considered here.

Finally, each iteration consists of $R$ rollouts with mixed data, depending on the number of robots running in parallel within the simulation, which are then used by the PPO algorithm to update the policy.
\section{Experimental Results}

This section details the experimental framework and evaluates the performance of the proposed MBRL controller against vanilla PPO, since the MBRL methods referenced in this paper assume an off-policy setting and therefore do not provide a fair comparison. We first conduct an ablative study to analyze how rollout-generation parameters influence training stability, providing a technical justification for our selected data augmentation configuration. Subsequently, we benchmark the framework across diverse legged platforms, evaluating key training metrics including cumulative reward and policy divergence. Finally, we provide a robust validation of the learned locomotion policy by subjecting the robot to a diverse set of velocity and directional commands.

\subsection{Platform}

Training is conducted within the Isaac Lab~\cite{mittal2025isaac} simulation environment, leveraging its high-fidelity physics engine and specialized reinforcement learning utilities. We utilize standard locomotion tasks for diverse quadrupedal platforms, including the ANYmal-C, ANYmal-D, Unitree Go1, and Go2 as benchmarks. To ensure an unbiased evaluation, our method is compared against vanilla PPO baselines using these standard task formulations. The proposed framework is integrated into the RSL-RL library~\cite{schwarke2025rsl}.

While the Unitree Go1 serves as the primary platform due to its extensive suite of established locomotion controllers, we also demonstrate training across additional robotic platforms. This cross-platform validation highlights that our hyperparameters are fundamentally coupled to the locomotion task dynamics rather than being overfitted to specific robotic hardware or reward weight configurations.

\subsection{Hardware}

To conduct the experiments, we rely on the Delta Cluster~\cite{boerner2023access}, allocated at the University of Illinois at Urbana-Champaign (UIUC) using an NVIDIA A40 GPU with a total of 48 GB VRAM. The high computational power of this GPU enables a deeper analysis of what could enhance the efficiency in the case of a high-end GPU, as well as in cases with restricted hardware.

\subsection{Ablative Study}
\label{sec:ablative_study}

Given that this work aims to enhance the sample efficiency of RL-based locomotion controllers, we conducted an ablation study on the baseline to characterize the performance trade-offs. This study investigates how the number of parallel environments and the rollout horizon influence training efficiency, specifically evaluating max reward, wall-clock time, and the deviation of the policy. We monitor the policy's deviation to assess convergence, as a lower variance indicates a policy that is more certain in its actions and has effectively narrowed down its decision-making distribution.

It is important to highlight that the number of parallel environments and the rollout horizon directly influence the amount of data that the PPO will face, but with different perspectives, as the algorithm evaluates the trajectory following the estimated advantage per step.

These results establish a performance ceiling for model-free methods and identify the specific configurations that stand to benefit most from rollout augmentation via synthetic transitions, as proposed in our framework.
 
We used the default Unitree Go1 flat-terrain task (\texttt{Isaac-Velocity-Flat-Unitree-Go1-v0}), maintaining a consistent policy architecture and hyperparameters (Appendix~\ref {app:ppo_settings}) across all trials. Using this environment, we conducted a systematic ablation study to analyze the sensitivity of the learning process to the number of parallel environments and the rollout horizon. We performed three independent training runs for each setting for 500 iterations, increasing and decreasing the standard parameters $R=1024$ and $N=24$, as detailed in Table~\ref{tab:ppo_ablation}.

From this data, it is possible to see how the rollout length impacts the learning time more significantly, as the data cannot be parallelized because the steps are sequence-dependent. Furthermore, a strong inverse correlation exists between policy standard deviation and the max reward; as the policy converges and confidence increases, indicated by lower deviation, the agent produces actions that align with the task objectives, improving the return.

\begin{table}[b]
\centering
\caption{Ablation study: Identifying the efficiency ceiling for baselines. \\
\tiny{* Max Reward denotes the average of the maximum reward across all seeds.}}
\label{tab:ppo_ablation}
\resizebox{\columnwidth}{!}{%
\begin{tabular}{l l r r r r r r}
\toprule
{Envs ($R$)} & {Metric} & {$N$=12} & {$N$=16} & {$N$=20} & {$N$=24} & {$N$=28} & {$N$=32} \\
\midrule
\multirow{3}{*}{512}  & Time (min)     & 3.1   & 3.8   & 4.6   & 5.3   & 6.0   & 6.8   \\
                      & Policy Std     & 0.63  & 0.69  & 0.65  & 0.68  & 0.69  & 0.69  \\
                      & Max Reward & 5.5   & 6.4   & 9.0   & 8.3   & 10.1  & 11.2  \\
\midrule
\multirow{3}{*}{1024} & Time (min)     & 3.2   & 4.0   & 4.8   & 5.6   & 6.3   & 7.2   \\
                      & Policy Std     & 0.64  & 0.64  & 0.65  & 0.67  & 0.59  & 0.48  \\
                      & Max Reward & 7.2   & 10.0  & 15.7  & 17.3  & 27.8  & 32.1  \\
\midrule
\multirow{3}{*}{2048} & Time (min)     & 3.4   & 4.3   & 5.3   & 6.0   & 7.0   & 7.8   \\
                      & Policy Std     & 0.61  & 0.60  & 0.53  & 0.39  & 0.41  & 0.42  \\
                      & Max Reward & 10.3  & 19.7  & 27.5  & 35.5  & 35.4  & 35.0  \\
\midrule
\multirow{3}{*}{4096} & Time (min)     & 3.9   & 4.9   & 5.8   & 6.8   & 7.9   & 9.0   \\
                      & Policy Std     & 0.63  & 0.42  & 0.37  & 0.38  & 0.37  & 0.38  \\
                      & Max Reward & 12.9  & 33.3  & 36.2  & 36.8  & 37.4  & 36.8  \\
\midrule
\multirow{3}{*}{8192} & Time (min)     & 5.1   & 6.3   & 7.8   & 9.3   & 10.8  & 12.3  \\
                      & Policy Std     & 0.43  & 0.34  & 0.36  & 0.36  & 0.36  & 0.37  \\
                      & Max Reward & 32.2  & 36.7  & 37.3  & 37.6  & 37.8  & 38.2  \\
\bottomrule
\end{tabular}%
}
\end{table}

\begin{figure}[t]
    \centering
    \includegraphics[width=\linewidth]{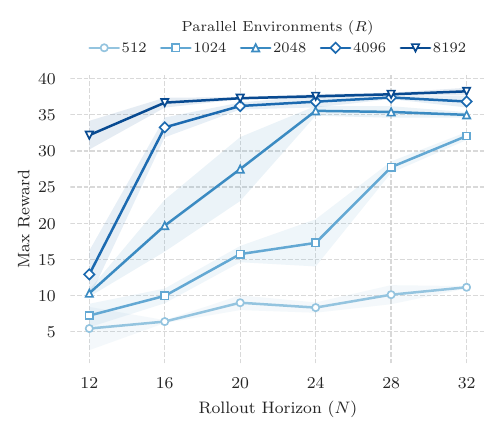}
    \caption{Performance sensitivity to data throughput parameters. The figure illustrates the maximum reward achieved after 500 iterations across varying environment counts and rollout horizons.}
    \label{fig:ablative_study}
\end{figure}

Evaluating specifically the max reward regarding the data parameters, as illustrated in Fig.~\ref{fig:ablative_study}, there is a clear correlation between environment scaling and the efficacy of the rollout horizon. At lower environment counts ($R=512$), the training performance remains degraded regardless of the rollout length, suggesting that a minimum level of sample diversity is required for stable convergence. Conversely, at $R=8192$, the policy consistently achieves peak reward values across all tested horizons. However, as already discussed, this high-throughput configuration entails a significant increase in wall-clock time.

In parallel, it is possible to identify a critical point for $R=4096$, where extending the rollout horizon from 12 to 16 yields the most substantial performance gain. This suggests that by augmenting shorter, computationally cheaper horizons with synthetic steps, we can achieve high-performance locomotion without simulating a massive number of parallel agents. Consequently, the configuration of 4096 environments with a rollout length of 16 is selected as the baseline for all subsequent experiments, serving as the foundation for evaluating our model-based data augmentation framework.

\subsection{Model Evaluation}

Before integrating the model directly into the data augmentation pipeline, we evaluated the training process using the parameters identified in the previous section. Our objective was to observe the model's accuracy and loss when trained on dynamic data provided by the policy while the policy itself is undergoing training. Through this empirical process, we defined the transition model settings used in subsequent steps, as detailed in Table~\ref{tab:predictive_model}.

\begin{table}[t]
\centering
\caption{Predictive model configuration.}
\begin{tabular}{cc}
\hline
Predictive Network      & MLP{[}256, 256, 256, 256{]} \\ \hline
Learning Rate           & 3e-3 \\ \hline
\# Epochs per Iteration & 25 \\ 
\hline
\end{tabular}
\label{tab:predictive_model}
\end{table}

Following the literature, we utilized a compact architecture to ensure the model adapts quickly to shifting policy distributions while remaining computationally efficient for data generation~\cite{janner2019trust}.

\begin{figure}[h]
    \centering
    \includegraphics[width=\linewidth]{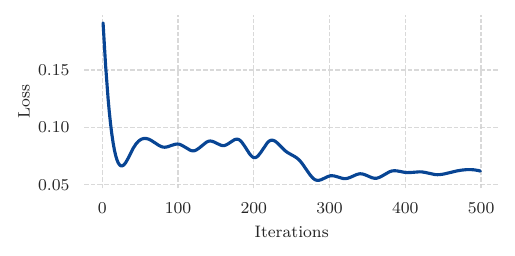}
    \caption{Predictive model convergence over training.}
    \label{fig:model_loss}
\end{figure}

Figure~\ref{fig:model_loss} illustrates the model's Huber loss during training. The significant initial drop represents a convergence phase that justifies delaying augmentation via our scheduler. In later stages, the loss remains nearly constant, demonstrating the model's ability to align with the evolving policy data distribution after the warmup period.

\begin{figure}[h]
    \centering
    \includegraphics[width=\linewidth]{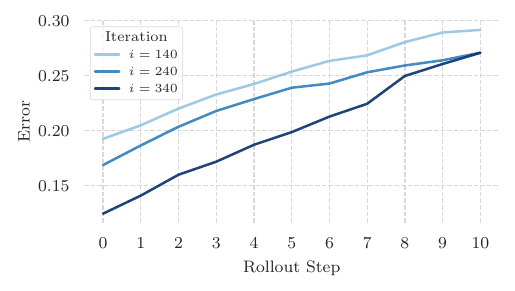}
    \caption{Evolution of the prediction horizon error using multi-step prediction.}
    \label{fig:horizon_eval}
\end{figure}

As our framework performs multi-step prediction, we evaluated the evolution of compounding error (Fig.~\ref{fig:horizon_eval}). While this error reduces after the warmup iterations, it scales with the prediction horizon. This accumulation of error provides a primary justification for utilizing short-horizon predictions during data augmentation.

An additional constraint necessitates short horizons; our model predicts rewards and observations but does not account for termination signals, such as collisions. This lack of termination prediction requires restricted horizons to maintain the physical integrity of the generated trajectories.

\subsection{Policy Training}

Utilizing the same policy architecture and hyperparameters established in previous sections, we evaluate the proposed Dyna-style method against the baseline PPO implementation. This comparison is conducted across the five distinct configurations detailed in Table~\ref{tab:training_configs}, each evaluated over three independent training runs for 500 iterations. Table~\ref{tab:step_time} reports the per-phase cost of a single transition step.

Furthermore, we extended the evaluation from the Unitree Go1 to the Unitree Go2, ANYmal-C, and ANYmal-D using identical prediction model hyperparameters. These results demonstrate that the methodology is scalable when evaluating the task rather than the specific platform and that synthetic data is beneficial for training. However, without platform-specific tuning, these gains do not consistently translate into improvements across all metrics, such as statistical significance, which is why we focus the analysis on the tuned platform for a more specific evaluation. This limitation arises in part from differences in robot morphology across platforms, which introduce varying modeling complexity for the learned dynamics, requiring the model to balance bias and variance during training~\cite{yang2020rethinking}. Moreover, because the framework splits observations into predictable and non-predictable components, the formulation must be adapted for platforms with very different morphologies, such as humanoids.

For the Unitree Go1 results in Table~\ref{tab:cross_platform_results}, the Vanilla PPO-12 baseline reaches only 12.96 reward, whereas 16 steps raise it to 33.27 at 25.64\% more wall-clock time.

\begin{table}[t]
\centering
\caption{Configurations used for policy training experiments.}
\begin{tabular}{cccc}
\hline
Approach & $N$ & $a \rightarrow b$ & $x \rightarrow y$ \\ 
\hline
Vanilla PPO & 12 & -- & -- \\
Vanilla PPO & 16 & -- & -- \\
Ours w/o scheduler (2-step) & 16  & $0 \rightarrow 0$ & $2 \rightarrow 2$ \\ 
Ours w/o scheduler (4-step)             & 16  & $0 \rightarrow 0$ & $4 \rightarrow 4$ \\
Ours (4-step)               & 16  & $0 \rightarrow 100$ & $0 \rightarrow 4$ \\ 
\hline
\end{tabular}
\label{tab:training_configs}
\end{table}

\begin{table*}[b]
\setcounter{table}{4}
\centering
\caption{Training performance across quadrupedal platforms (mean $\pm$ standard deviation over three seeds).}
\label{tab:cross_platform_results}
\resizebox{0.95\textwidth}{!}{%
\begin{tabular}{cc cccc c c}
\toprule
\multirow{2}{*}{Platform} & \multirow{2}{*}{Approach} & \multicolumn{4}{c}{Time (min)} & \multirow{2}{*}{Policy Std} & \multirow{2}{*}{Max Reward} \\
\cmidrule(lr){3-6}
& & Sim. Collection & Synthetic Gen. & Policy Learning & {Total} & & \\
\midrule
\multirow{5}{*}{\shortstack{Unitree Go1 \\ \texttt{Isaac-Velocity-Flat-Unitree-Go1-v0}}} & Vanilla PPO (12-step)      & 3.1 & --  & 0.8 & 3.9 & $0.629 \pm 0.013$ & $12.96 \pm 2.68$ \\
                               & Vanilla PPO (16-step)      & 4.1 & --  & 0.8 & 4.9 & $0.419 \pm 0.029$ & $33.27 \pm 1.76$ \\
                               & Ours w/o Sch (2-step)      & 3.5 & 0.5 & 0.8 & 4.8 & $0.441 \pm 0.015$ & $33.42 \pm 0.78$ \\
                               & Ours w/o Sch (4-step)      & 3.1 & 0.4 & 0.8 & 4.3 & $0.612 \pm 0.040$ & $\phantom{0}9.17 \pm 1.48$ \\
                               & Ours (4-step)              & 3.1 & 0.4 & 0.8 & 4.3 & $0.447 \pm 0.009$ & $33.35 \pm 0.54$ \\
\midrule
\multirow{5}{*}{\shortstack{Unitree Go2 \\ \texttt{Isaac-Velocity-Flat-Unitree-Go2-v0}}} & Vanilla PPO (12-step)      & 3.2 & --  & 0.8 & 4.0 & $0.437 \pm 0.060$ & $25.07 \pm 5.38$ \\
                               & Vanilla PPO (16-step)      & 4.0 & --  & 0.8 & 4.8 & $0.330 \pm 0.004$ & $33.01 \pm 0.86$ \\
                               & Ours w/o Sch (2-step)      & 3.4 & 0.5 & 0.8 & 4.7 & $0.359 \pm 0.001$ & $34.51 \pm 0.12$ \\
                               & Ours w/o Sch (4-step)      & 3.0 & 0.4 & 0.8 & 4.2 & $0.402 \pm 0.012$ & $33.12 \pm 0.64$ \\
                               & Ours (4-step)              & 3.1 & 0.4 & 0.8 & 4.3 & $0.373 \pm 0.007$ & $34.88 \pm 0.69$ \\
\midrule
\multirow{5}{*}{\shortstack{ANYmal-C \\ \texttt{Isaac-Velocity-Flat-Anymal-C-v0}}}      & Vanilla PPO (12-step)      & 3.3 & --  & 0.8 & 4.1 & $0.265 \pm 0.009$ & $10.75 \pm 5.94$ \\
                               & Vanilla PPO (16-step)      & 4.5 & --  & 0.8 & 5.3 & $0.255 \pm 0.008$ & $18.31 \pm 0.65$ \\
                               & Ours w/o Sch (2-step)      & 3.7 & 0.5 & 0.8 & 5.0 & $0.257 \pm 0.011$ & $19.35 \pm 1.13$ \\
                               & Ours w/o Sch (4-step)      & 3.2 & 0.5 & 0.8 & 4.5 & $0.261 \pm 0.024$ & $11.12 \pm 1.34$ \\
                               & Ours (4-step)              & 3.4 & 0.4 & 0.8 & 4.6 & $0.256 \pm 0.013$ & $16.80 \pm 2.48$ \\
\midrule
\multirow{5}{*}{\shortstack{ANYmal-D \\ \texttt{Isaac-Velocity-Flat-Anymal-D-v0}}}      & Vanilla PPO (12-step)      & 3.5 & --  & 0.8 & 4.3 & $0.314 \pm 0.061$ & $\phantom{0}6.85 \pm 3.18$  \\
                               & Vanilla PPO (16-step)      & 4.6 & --  & 0.8 & 5.4 & $0.280 \pm 0.018$ & $12.61 \pm 5.86$ \\
                               & Ours w/o Sch (2-step)      & 4.2 & 0.5 & 0.8 & 5.5 & $0.303 \pm 0.011$ & $12.64 \pm 6.17$ \\
                               & Ours w/o Sch (4-step)      & 3.5 & 0.4 & 0.8 & 4.8 & $0.325 \pm 0.013$ & $\phantom{0}7.30 \pm 2.32$ \\
                               & Ours (4-step)              & 3.6 & 0.4 & 0.8 & 4.8 & $0.290 \pm 0.004$ & $11.61 \pm 3.31$ \\
\bottomrule
\end{tabular}%
}
\end{table*}

\begin{table}[H]
\setcounter{table}{3}
\centering
\caption{Average time to compute synthetic steps and update the model.\\
\tiny{* Results obtained with 4,096 parallel environments.}}
\begin{tabular}{l c}
\hline
Phase & Avg. Time (ms) \\
\hline
Ground-truth Simulation ($p$)  & 31 per Step \\
Synthetic Generation ($\hat{p}$)  & \phantom{0}2 per Step \\
Model Training ($\mathcal{L}^{\text{model}}$)  & 45 \\
Policy Training ($\mathcal{L}^{\text{CLIP}}$)  & 97 \\
\hline
\end{tabular}
\label{tab:step_time}
\end{table}

Our proposed scheduled 4-step method bridges this gap by achieving a comparable reward of 33.35 using the same amount of simulated data as the 12-step baseline. Across three independent runs on the tuned Go1, a Welch’s $t$-test confirms this reward improvement is statistically significant (p = $0.0044$), effectively providing the benefits of a longer horizon while reducing the total training time by 12.24\% compared to the full 16-step simulation. Notably, our method also yields a lower standard deviation, suggesting that the shifted transition model distribution enhances exploration stability.

Evaluation of alternative configurations validates the scheduling strategy. The unscheduled 2-step variant keeps compounding model error small, tolerating a lower-accuracy model, but its small synthetic proportion yields only marginal wall-clock savings. The 4-step setting offers higher efficiency potential, but demands a more accurate model over the longer horizon; without the scheduler, inaccurate early-stage predictions pollute the buffer and the policy collapses. The scheduler unlocks the 4-step regime, but realizing it requires per-platform tuning, as the cross-platform results show.

Table~\ref{tab:step_time} further summarizes the average time spent on physics simulation, model training, and synthetic step inference, providing a clearer breakdown of our method’s efficiency in wall-clock terms. Although a synthetic step costs only 6.4\% of a physics step (2 ms vs. 31 ms), policy training and model training are fixed per-iteration costs that substitution cannot reduce, which is why the wall-clock saving is smaller than the reduction in simulation steps on a task this inexpensive to simulate.

Analysis across the remaining platforms allows us to evaluate whether the framework generalizes to similar tasks that differ only in reward functions and quadruped morphology. Across these tasks, our method with a 4-step pipeline and scheduling consistently outperforms the Vanilla PPO-12 baseline and provides notable wall-clock advantages compared to running Vanilla PPO-16. 

However, the outcome differs by morphology. On the Unitree Go2, which shares the Go1's morphology, our method matches or exceeds the Vanilla PPO-16 baseline, whereas on the morphologically distinct ANYmal-C and ANYmal-D platforms, it approaches but does not surpass the longer baseline, likely due to the need for per-platform tuning of the model hyperparameters.

\begin{table}[h]
\setcounter{table}{5}
\centering
\caption{Training efficiency and convergence metrics ($\delta$ = 30.5).}
\label{tab:go1_efficiency}
\begin{tabular}{lccc}
\toprule
Method & SR to $\delta$ (\%) & Sim. Steps to $\delta$ (M) & Iter. \\
\midrule
Vanilla PPO (12-step) & 0\% & --- & --- \\
Vanilla PPO (16-step) & 100\% & 27.53 & 420 \\
Ours w/o Sch (2-step)  & 100\% & 22.82 & 398 \\
Ours w/o Sch (4-step)  & 0\% & --- & --- \\
\textbf{Ours (4-step)} & \textbf{100\%} & \textbf{19.64} & \textbf{383} \\
\bottomrule
\end{tabular}
\flushleft
\scriptsize {\textbf{Note:} $\delta$  is defined as 80\% of the mean peak reward achieved by the experiments described in Section~\ref{sec:ablative_study}. SR denotes success rate.}
\end{table}

To evaluate the task used for hyperparameter tuning (Unitree Go1), Table~\ref{tab:go1_efficiency} reports the simulation steps required to reach $\delta$, a convergence threshold defined as 80\% of the mean peak reward from the experiments in Section~\ref{sec:ablative_study}. This metric evaluates both sample efficiency and reward progression. While the baseline policy without rollout extensions failed to reach this threshold, our extended approach successfully attained $\delta$ using fewer simulation steps than the 16-step baseline (19.64M vs. 27.53M) and requiring fewer training iterations. This confirms that the model-generated experience not only maintains performance but also enhances computational efficiency, as these iterations are faster to compute.

\begin{figure}[t]
    \centering
    \includegraphics[width=\linewidth]{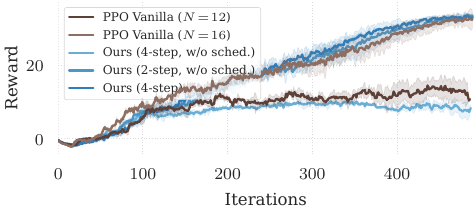}
    \caption{Evolution of reward across training iterations for all evaluated settings. In all evaluated variants of our proposed method, $N=16$.}
    \label{fig:learning_curve}
\end{figure}

The training curves for the Unitree Go1 in Fig.~\ref{fig:learning_curve} illustrate the progression toward maximum reward and learning variance, providing context for the wall-clock and reward performance data in Table~\ref{tab:cross_platform_results}. Notably, our 2-step and 4-step methods replicate the learning dynamics of the Vanilla-16 baseline while significantly reducing training time per iteration. In contrast, Vanilla-12 fails to sufficiently explore the observation space, resulting in plateaued rewards. Similarly, the 4-step method without scheduling suffers from early-stage exploration issues, where noisy updates prematurely guide the policy toward suboptimal behaviors.

\begin{figure}[t]
    \centering
    \includegraphics[width=\linewidth]{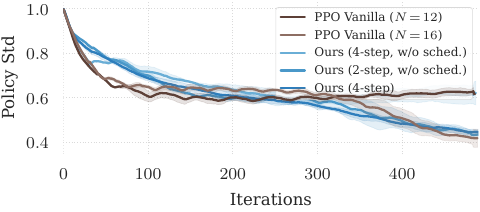}
    \caption{Evolution of policy deviation across training iterations for all evaluated settings. In all evaluated variants of our proposed method, $N=16$.}
    \label{fig:policy_curve}
\end{figure}

Additionally, we assessed the quality of the learned actions by tracking the distributional confidence of the policy, as illustrated in Fig.~\ref{fig:policy_curve}. Since the reward function already captures the robot’s tracking error and significantly influences the overall return, we further analyzed the distributional confidence to determine whether high tracking performance results from precise decision-making or from uncertainty in action selection. The policy deviation under our approach closely aligns with that of standard Vanilla PPO-16, suggesting that the entropy of the action distribution decreases at a robust rate. This indicates that the resulting policy is not only high-performing but also stable and reliable, effectively following complex commands while preserving the natural gait induced by the auxiliary reward terms. In addition, Appendix~\ref{app:kl_div} discusses the potential policy shift during training caused by the synthetic data.

It is worth noting that achieving high rewards early in training establishes a strong foundation for subsequent learning. As the return approaches its maximum, the policy can then focus on minimizing variance, fostering the stable and repeatable behaviors required for real-world robot locomotion. While the examples presented here converge within minutes, the method is designed to scale to more complex proprioceptive (blind) tasks, such as~\cite{zhu2025robust, angarola2025learning, wu2026toward}, that typically require several hours or even days of training.

This approach provides a means to maintain stability over long horizons while improving time efficiency and reducing the variability of final rewards across independent runs.

\subsection{Policy Evaluation}
\label{sec:policy_evaluation}

\begin{figure*}[t]
    \centering
    \includegraphics[width=0.82\linewidth]{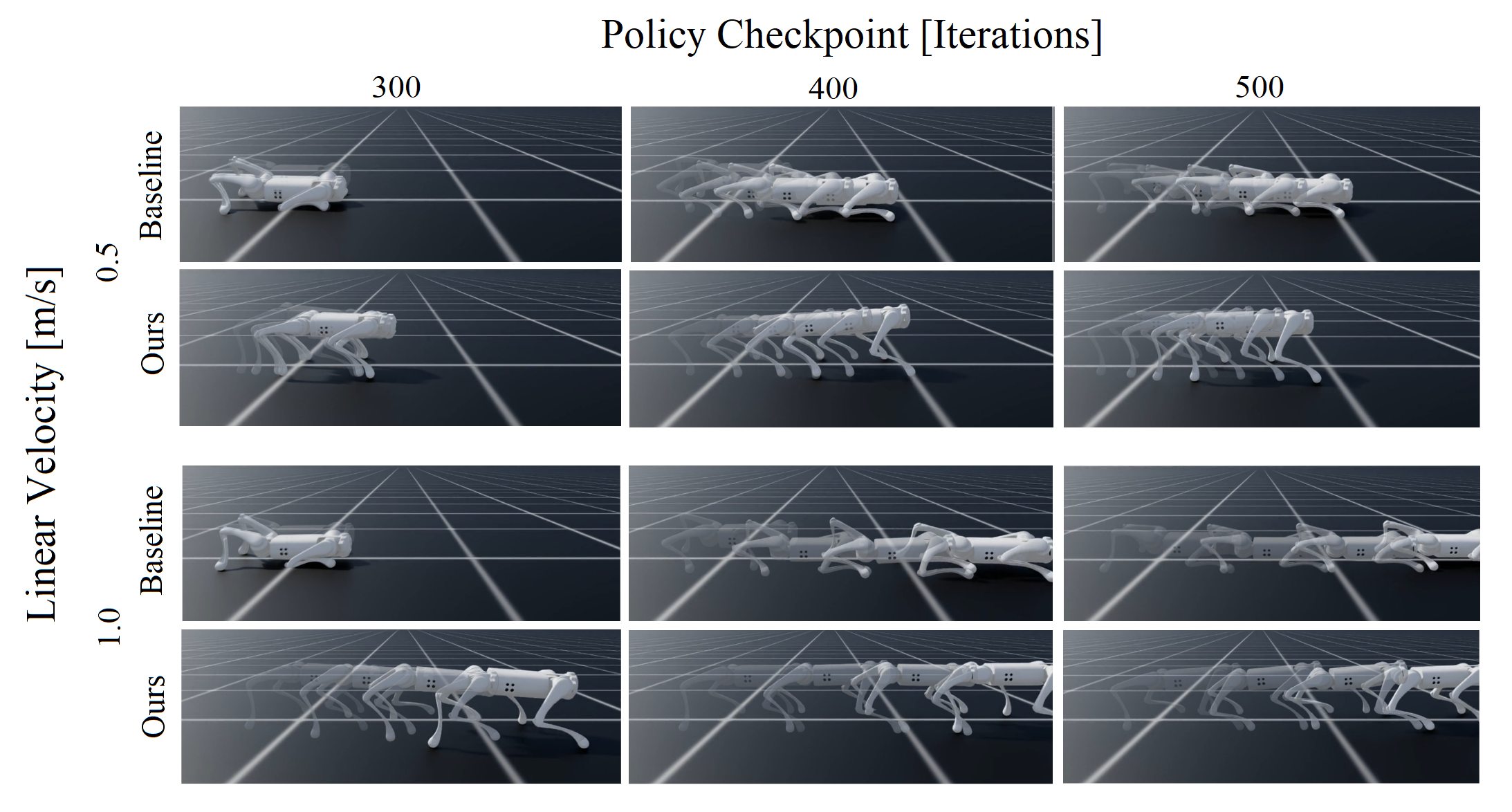}
    \caption{Qualitative comparison of locomotion emergence across training checkpoints. Each comparison between the PPO-16 baseline and the proposed 4-step method across three training checkpoints (300, 400, 500 iterations) using identical seeds. While the baseline fails to produce effective locomotion at early stages, our method exhibits clear gait emergence and better adherence to regularization terms (e.g., base orientation and air time) at both 0.5~m/s and 1.0~m/s command velocities.}
    \label{fig:policy_evaluation}
\end{figure*}

While the previous subsection established the advantages of the proposed pipeline through reward, convergence, and wall-clock metrics, we now evaluate the quality of the learned locomotion. This analysis demonstrates that, beyond improving training efficiency, our pipeline enables the robot to acquire a stable walking gait significantly earlier than the baseline PPO. To illustrate this, Fig.~\ref{fig:policy_evaluation} compares the motion generated by both methods using policies from checkpoints 300, 400, and 500. To ensure a fair comparison, we selected policies trained under the same seed and evaluated them at commanded linear velocities of 0.5~m/s and 1.0~m/s.

At iteration 300, the baseline policy fails to produce effective forward locomotion for either command. In contrast, the policy trained with our pipeline already exhibits clear stepping and visible progression. This distinction is crucial; it demonstrates that our method does not merely improve final training statistics but significantly accelerates the emergence of functional locomotion by balancing movement rewards with style and regularization terms.

In the final stages of training, our method continues to improve numerically; however, the primary evolution occurs in the robot's behavior, which moves toward a lower-variance gait that strictly adheres to the style constraints. Specifically, the policy increasingly adheres to regularization terms, such as flat base orientation and foot air time (Appendix~\ref{app:rewards}), resulting in a more refined locomotion style. Conversely, the baseline continues to struggle with these regularization terms. Because velocity tracking represents the highest magnitude component of the reward function, the baseline prioritizes command following at the expense of motion quality.

These observations align with the quantitative data in Table~\ref{tab:go1_efficiency}. The scheduled 4-step method achieves rewards comparable to PPO-16 in fewer simulated steps and reaches the convergence threshold $\delta$ with fewer iterations and simulation steps. Similarly, the learning curves (Fig.~\ref{fig:learning_curve}) and policy deviation trends (Fig.~\ref{fig:policy_curve}) show that our method reaches high-reward behavior earlier with more stable refinement. Collectively, these results demonstrate that the proposed pipeline improves both training efficiency and the practical emergence of locomotion, enabling earlier acquisition of high-quality gaits.
\section{Conclusion and Future Work}

This work evaluates the integration of Dyna-style model-based reinforcement learning into a PPO-based framework to improve the training efficiency of quadrupedal locomotion controllers. By progressively substituting trajectory tails with short-horizon synthetic data from a learned transition model, the method reduces the reliance on computationally intensive physics-based simulations. A key factor in maintaining the stability of this approach is the ground-truth anchoring mechanism, which re-initializes each new rollout iteration from a verified simulation state, thereby preventing the compounding errors inherent in predictive models from destabilizing the policy. Experimental results across multiple platforms show that this framework reduces wall-clock training time by 12.24\% on the tuned Go1 platform while matching or exceeding the rewards of model-free PPO baselines on the Unitree Go1 and Go2 platforms, and approaching their performance on the untuned ANYmal platforms.

This work is restricted by the need to re-train the transition model every iteration to stay synchronized with the evolving policy. Beyond the blind locomotion setting studied here, richer world models predicting both proprioceptive states and exteroceptive features could extend the framework to perception-aware locomotion. Finally, although the framework is designed to scale to more computationally expensive, long-horizon tasks, all tasks evaluated here converge within minutes and assume no termination signals (e.g., collisions). Extending it to more complex MDPs may require additional mechanisms to account for such events, an important direction for future work.

\appendices

\section{Environment Settings}
\label{app:additional_params}

This appendix provides detailed implementation notes for the training environments used in this work. It should be noted that all tasks are defined and available within Isaac Lab~\cite{mittal2025isaac}.

\subsection{Rewards}
\label{app:rewards}

Following Section~\ref{sec:rewards}, the locomotion reward is defined as the weighted sum of a task component, which enforces command tracking, and a style component, which regularizes the motion and discourages undesirable behaviors:
\begin{equation}
r(o_k,a_k) = r^{\text{task}}_k + r^{\text{style}}_k
= \sum_{i} w_i  r_i(o_k,a_k),
\label{eq:reward_sum}
\end{equation}
where $w_i$ is the scalar weight of reward term $r_i(\cdot)$.

In the evaluated tasks, $r^{\text{task}}$ is composed of exponential tracking terms for base linear velocity in the horizontal plane and base yaw rate. The remaining terms form $r^{\text{style}}$ and act as regularizers (e.g., discouraging vertical motion, roll/pitch oscillations, excessive torques or accelerations, and abrupt action changes) and, when enabled, contact-safety violations. For completeness and reproducibility, Table~\ref{tab:reward_terms_all_tasks} reports the reward definitions and weights.

\begin{table*}[b]
\centering
\footnotesize
\setlength{\tabcolsep}{3pt}
\renewcommand{\arraystretch}{1.2}
\caption{Reward and penalty terms for velocity-tracking tasks. A dash (--) indicates the term is inactive for that specific robot family.}
\label{tab:reward_terms_all_tasks}
\begin{tabular}{p{0.25\textwidth} p{0.10\textwidth} p{0.45\textwidth} r r}
\hline
\textbf{Reward Component} & \textbf{Symbol} & \textbf{Mathematical Formulation} & \textbf{ANYmal} & \textbf{Go1/Go2} \\
\hline
\textit{Task Rewards} & & & & \\
Linear Velocity Tracking & $r_{v_{x,y}}$ & $\exp(-\|v_{xy} - v^{\text{cmd}}_{xy}\|_2^2)$ & 1.0 & 1.5 \\
Angular Velocity Tracking & $r_{\omega_z}$ & $\exp(-(\omega_z - \omega^{\text{cmd}}_{z})^2)$ & 0.5 & 0.75 \\
\hline
\textit{Style Rewards} & & & & \\
Base Motion (Vertical) & $r_{v_z}$ & $v_z^2$ & -2.0 & -2.0 \\
Base Motion (Angular) & $r_{\omega_{xy}}$ & $\|\omega_{xy}\|_2^2$ & -0.05 & -0.05 \\
Flat Orientation & $r_{\text{ori}}$ & $\|g_{xy}\|_2^2$ & -5.0 & -2.5 \\
Joint Torque & $r_{\tau}$ & $\|\tau\|_2^2$ & -2.5e-5 & -2e-4 \\
Joint Acceleration & $r_{\ddot{q}}$ & $\|\ddot{q}\|_2^2$ & -2.5e-7 & -2.5e-7 \\
Action Rate & $r_{\Delta a}$ & $\|a_k - a_{k-1}\|_2^2$ & -0.01 & -0.01 \\
Undesired Contacts & $r_{\text{coll}}$ & $\sum_{\text{non-foot}} \mathds{1}_{\text{contact}}$ & -1.0 & -- \\
Feet Air Time & $r_{\text{air}}$ & $\sum_{\text{feet}} \mathds{1}_{\text{air}}$ & 0.5 & 0.25 \\
\hline
\end{tabular}

\vspace{4pt}
\begin{flushleft}
\scriptsize
\textbf{Note:} All terms are computed using proprioceptive feedback and current setpoints. $\mathds{1}_{\text{air}}$ and $\mathds{1}_{\text{contact}}$ represent indicator functions for the presence of air time and undesired contact, respectively. $\tau \in \mathbb{R}^{12}$ refers to the joint torques.
\end{flushleft}
\end{table*}

\subsection{Terminations}
\label{app:terminations}

All environments in the benchmark share a unified termination logic governed by two primary conditions. First, episodes are truncated upon reaching a fixed horizon of $T$ timesteps (where $k \geq T$); this is flagged as a "time-out" and treated as a standard truncation rather than a failure. Second, an early termination is triggered if the robot’s base or non-pedal components contact the ground. This failure criterion penalizes instability and prevents the policy from exploiting idiosyncratic or non-upright behaviors.

\subsection{Command Sampling}
\label{app:commands}

The velocity-tracking tasks are command-conditioned. Specifically, a command vector is sampled for each robot according to the uniform distributions defined in equation~(\ref{eq:command_uniform}).  
\begin{align}
    \label{eq:command_uniform}
    v^\text{cmd}_{x} &\sim \mathcal{U}\left(v^{\min}_{x}, v^{\max}_{x}\right), \nonumber\\
    v^{\text{cmd}}_{y} &\sim \mathcal{U}\left(v^{\min}_{y}, v^{\max}_{y}\right), \nonumber\\
    \omega^{\text{cmd}}_{z} &\sim \mathcal{U}\left(\omega^{\min}_{z}, \omega^{\max}_{z}\right).
\end{align}

In our experimental setup, the linear velocity and yaw rate command ranges are restricted to $v_{x,y}^{\mathrm{cmd}} \in [-1, 1]~\text{m/s}$ and $\omega_z^{\mathrm{cmd}} \in [-1, 1]~\text{rad/s}$, respectively. The sampled command remains constant throughout the duration of the episode. Upon an episode reset, a new command is sampled.

\section{PPO hyperparameters}
\label{app:ppo_settings}

This section summarizes the PPO hyperparameters and policy network settings used in our experiments. Since the configurations are identical within each robot family, Table~\ref{tab:ppo_settings} reports the values for ANYmal-C/D and Go1/Go2.

\begin{table}[H]
\centering
\footnotesize
\renewcommand{\arraystretch}{1.1}
\caption{PPO and network hyperparameters.}
\label{tab:ppo_settings}
\begin{tabular}{l c}
\hline
\textbf{Hyperparameter} & \textbf{Value} \\
\hline
Actor MLP hidden sizes & MLP[128, 128, 128] \\
Critic MLP hidden sizes & MLP[128, 128, 128] \\
Discount factor & 0.99 \\
GAE parameter & 0.95 \\
PPO clipping parameter & 0.2 \\
Value loss coefficient & 1.0 \\
Entropy coefficient & [0.005, 0.01] \\
Learning epochs per update & 5 \\
Mini-batches per epoch & 4 \\
Target KL divergence & 0.01 \\
Max gradient norm & 1.0 \\
\hline
\end{tabular}
\vspace{2pt}
\begin{flushleft}
\scriptsize
\textbf{Note:} The entropy coefficient is the only parameter that differs between the ANYmal and Unitree platforms. For entries represented as a vector $[x, y]$, the first value refers to ANYmal and the second to Unitree.
\end{flushleft}
\end{table}

\subsection{KL-Divergence-Based Learning Rate Adaptation}
\label{app:kl_div}

Because augmenting PPO on-policy trajectories with synthetic data through the approach presented in this paper may violate the algorithm's core assumptions, we discuss the potential effects of policy shift due to this additional imagined data. One standard approach to measure this shift is through the KL-divergence between policies during policy updates.

In this context, although PPO replaces explicit trust-region constraints like KL-divergence with the clipped objective function shown in equation~(\ref{eq:clip}), we utilize KL-divergence to adapt the learning rate as presented in~\cite{schulman2017proximal}. This approach uses the KL-divergence between the old and current policy, $\hat{\mathbb{E}_t} \left[ \text{KL} \left[ \pi_{\phi_{\text{old}}} (\cdot | s_t), \pi_{\phi} (\cdot | s_t) \right] \right]$, and compares it with the target value to decide if the learning rate will be increased or decreased, facilitating robust exploration during initial training phases while constraining policy updates in later stages to ensure stability.

\begin{figure}[t]
    \centering
    \includegraphics[width=\linewidth]{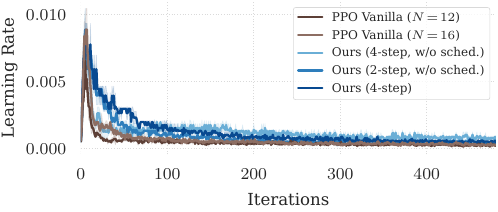}
    \caption{Evolution of the learning rate across training steps for all evaluated settings. In all evaluated variants of our proposed method, $N = 16$.}
    \label{fig:learning_rate}
\end{figure}

Acknowledging this direct relationship between the policy shift and the learning rate during training, Fig.~\ref{fig:learning_rate} provides qualitative supplementary results to evaluate the settings underlying Table~\ref{tab:training_configs}. The results suggest that the policy shift has an impact during early stages, but at convergence, the values are nearly aligned with the baselines, confirming the expected results due to the policy deviation convergence in Fig.~\ref{fig:policy_curve}.

\section*{Acknowledgment}
This work was supported in part by the São Paulo Research Foundation (FAPESP) under Grants no.~22/03339-8 and 23/17678-1, and by the Brazilian National Research Council (CNPq) under Grant no.~308092/2020-1. The authors also acknowledge the support of Petrobras, through resources from the ANP R\&D clause, in partnership with the University of São Paulo and the FAFQ foundation, under Cooperation Agreements no.~2023/00016-6 and 2023/00013-7. This work also used the Delta system through allocation CIS251247 from the Advanced Cyberinfrastructure Coordination Ecosystem: Services \& Support (ACCESS) program, which is supported by National Science Foundation Grants \#2138259, \#2138286, \#2138307, \#2137603, and \#2138296.

\bibliographystyle{IEEEtran}
\bibliography{ref}

\begin{IEEEbiographynophoto}{Francisco Affonso}
is currently pursuing the Ph.D. degree in computer science with the University of Illinois at Urbana-Champaign (UIUC), Urbana, IL, USA, where he is a Graduate Researcher with the Distributed Autonomous Systems Laboratory (DASLab). He received the B.S. degree in mechatronics engineering from the University of São Paulo (USP), São Carlos, Brazil. His research focuses on the development of loco-manipulation frameworks for robots operating in outdoor environments, emphasizing the integration of physics-based understanding and long-term reasoning for advanced task planning and action execution.
\end{IEEEbiographynophoto}

\begin{IEEEbiographynophoto}{Felipe Tommaselli}
is currently pursuing the Ph.D. degree in mechanical engineering, with an emphasis in mechatronics, with the São Carlos School of Engineering (EESC), University of São Paulo (USP), São Carlos, Brazil. He received the B.S. degree in electrical engineering, with a minor in computer science, from the same institution. His research interests include representation learning for scalable autonomous navigation.
\end{IEEEbiographynophoto}

\begin{IEEEbiographynophoto}{João H. Aléssio}
is currently pursuing the B.S. degree in mechatronics engineering with the São Carlos School of Engineering (EESC), University of São Paulo (USP), São Carlos, Brazil. He received the Mechatronics Technician degree from the Federal Institute of Santa Catarina (IFSC), Santa Catarina, Brazil. His research interests include autonomous mobile robotics, control systems, and world models.
\end{IEEEbiographynophoto}

\begin{IEEEbiographynophoto}{Mateus V. Gasparino}
is currently an Applied Scientist with Amazon Robotics. He received the B.S. degree in mechatronics engineering and the M.S. degree in mechanical engineering from the University of São Paulo (USP), São Carlos, Brazil, and the Ph.D. degree in computer science from the University of Illinois at Urbana-Champaign (UIUC), Urbana, IL, USA. His research interests include multi-modal perception and tracking, learning-based control systems, and the application of machine learning to robotics.
\end{IEEEbiographynophoto}

\begin{IEEEbiographynophoto}{Vivian Suzano Medeiros}
is currently a Postdoctoral Researcher with the Mobile Robotics Group, University of São Paulo (USP), São Carlos, Brazil. She received the B.S. degree in control and automation engineering and the M.S. and Ph.D. degrees in mechanical engineering from the Pontifical Catholic University of Rio de Janeiro (PUC-Rio), Rio de Janeiro, Brazil, in 2012, 2015, and 2020, respectively. During her doctorate, she spent one year as a Visiting Ph.D. Student with the Autonomous Systems Laboratory, Swiss Federal Institute of Technology, Zurich, Switzerland. At that time, she was involved in research on trajectory optimization and control for hybrid wheeled-legged robots. Her research interests include autonomous mobile robots, motion planning for legged and wheeled-legged robots, and control systems.

\end{IEEEbiographynophoto}

\begin{IEEEbiographynophoto}{Girish Chowdhary}
is currently an Associate Professor and the Donald Biggar Willett Faculty Fellow with the University of Illinois at Urbana-Champaign (UIUC), Urbana, IL, USA, where he is also the Director of the Field Robotics Engineering and Science Hub (FRESH) and the Director of the USDA/NIFA Farm of the Future. He holds a joint appointment with the Department of Agricultural and Biological Engineering and the Department of Computer Science, is a member of the Coordinated Science Laboratory, and holds affiliate appointments with Aerospace Engineering and Electrical Engineering. He received the Ph.D. degree in aerospace engineering from the Georgia Institute of Technology, Atlanta, GA, USA, in 2010. From 2003 to 2006, he was with the Institute of Flight Systems, German Aerospace Center (DLR), Braunschweig, Germany. From 2011 to 2013, he was a Postdoctoral Associate with the Laboratory for Information and Decision Systems (LIDS), Massachusetts Institute of Technology, Cambridge, MA, USA, and from 2013 to 2016, he was an Assistant Professor with Oklahoma State University, Stillwater, OK, USA. He is a recipient of the Air Force Young Investigator Award and several best paper awards, including the Best Systems Paper Award at RSS 2018 for his work on the agricultural robot TerraSentia.
\end{IEEEbiographynophoto}

\begin{IEEEbiographynophoto}{Marcelo Becker}
has been a Professor with the University of São Paulo (USP), São Carlos, Brazil, since 2008. He received the M.S. and Ph.D. degrees in mechanical engineering from the State University of Campinas (UNICAMP), Campinas, Brazil, in 1997 and 2000, respectively. During his Ph.D. studies, he spent eight months as a Guest Student with the Institute of Robotics, Swiss Federal Institute of Technology, Zurich, Switzerland, where he was involved in research on obstacle avoidance and map-building procedures for indoor mobile robots. From August 2005 to July 2006, he was on sabbatical leave with the Autonomous Systems Laboratory, Swiss Federal Institute of Technology, Lausanne, Switzerland, where he continued working on obstacle avoidance for indoor and outdoor mobile robots. His research interests include mobile robots, inspection robots, vehicular dynamics, design methodologies and tools, and mechanical design applied to robots and mechatronics.
\end{IEEEbiographynophoto}

\EOD

\end{document}